\documentclass{article}

\PassOptionsToPackage{numbers}{natbib}

\usepackage[table]{xcolor}

\usepackage[preprint]{neurips_2023}

\usepackage[utf8]{inputenc} 
\usepackage[T1]{fontenc}    
\usepackage{hyperref}       
\usepackage{url}            
\usepackage{booktabs}       
\usepackage{amsfonts}       
\usepackage{nicefrac}       
\usepackage{microtype}      
\usepackage{xcolor}         
\usepackage[pdftex]{graphicx}
\usepackage{array}
\usepackage[frozencache,cachedir=minted_frozen_cache]{minted}
\usepackage[most,minted]{tcolorbox}

\pdfoutput=1

\def\mund#1{\smallskip\noindent{\bf #1: }}
\newcommand{\mytcbinput}[4]{
\tcbinputlisting{
      listing file=#1,
      minted options={
        highlightlines={#3}, 
        highlightcolor=yellow,
        breaklines=True,        
        fontsize=\scriptsize,
        escapeinside=||,
      },
      breakable,
      title=#2,
      label=lst:#4,
      listing only
    }
}

\makeatletter
\newcolumntype{V}[1]{>{\topsep=0pt\@minipagetrue}p{#1}<{\vspace{-\baselineskip}}}
\makeatother

\usepackage[font=scriptsize,margin=1cm]{caption}
\usepackage{subcaption}

\title{GPT-4 Doesn't Know It's Wrong: An Analysis of Iterative Prompting for Reasoning Problems}

\author{%
  Kaya Stechly\thanks{Arizona State University, Tempe.} \hspace*{1cm} Matthew Marquez$^*$ \hspace*{1cm} Subbarao Kambhampati$^*$
}

\begin{document}

\maketitle

\begin{abstract}
There has been considerable divergence of opinion on the reasoning abilities of Large Language Models (LLMs). While the initial optimism that reasoning might emerge automatically with scale has been tempered thanks to a slew of counterexamples--ranging from multiplication to simple planning, there is still the wide spread belief that LLMs can self-critique and improve their own solutions in an iterative fashion. This belief seemingly rests on the assumption that verification of correctness should be easier than generation--a rather classical argument from computational complexity, that should be irrelevant to LLMs to the extent what they are doing is approximate retrieval. In this paper, we set out to systematically investigate the effectiveness of iterative prompting of LLMs in the context of {\em Graph Coloring}, a canonical NP-complete reasoning problem that is related to propositional satisfiability as well as practical problems like scheduling and allocation. We present a principled empirical study of the performance of GPT4 in solving graph coloring instances or verifying the correctness of candidate colorings--both in direct and iterative modes. In iterative modes, we experiment both with the model critiquing its own answers and an external correct reasoner verifying proposed solutions. 
In both cases, we analyze whether the content of the criticisms actually affects bottom line performance.
The study seems to indicate that (i) LLMs are bad at solving graph coloring instances (ii) they are no better at verifying a solution--and thus are not effective in iterative modes with LLMs critiquing LLM-generated solutions 
(iii) the correctness and content of the criticisms--whether by LLMs or external solvers--seems largely irrelevant to the performance of iterative prompting. 
We show that the observed effectiveness of LLMs in iterative settings is largely due to the correct solution being fortuitously present in the top-k completions of the prompt (and being recognized as such by an external verifier). Our results thus call into question claims about the self-critiquing capabilities of state of the art LLMs. 

\end{abstract}

\section{Introduction}

Large Language Models (LLMs), essentially n-gram models on steroids which have been trained on web-scale language corpus, have caught the imagination of the AI research community with linguistic behaviors that no one expected text completion systems to possess. Their seeming versatility has lead many researchers to wonder whether they can also do well on reasoning tasks typically associated with system 2 competency. Initial excitement based on anecdotal performance of LLMs on reasoning tasks has dissipated to some extent by the recent spate of studies questioning the robustness of such behaviors--be it planning \cite{valmeekam_large_2023,rao-cacm-blog-llm-reasoning}, simple arithmetic and logic \cite{dziri_faith_2023}, or general mathematical and abstract benchmark\cite{sawada_arb_2023,gendron2023large}. There still exists considerable optimism that even if LLMs can't generate correct solutions in one go, their accuracy improves in a iterative prompting regime, where LLMs will be able to "self-critique" their candidate solutions and refine them to the point of correctness \cite{yao2023react,tree-of-thoughts,shinn_reflexion_2023, weng2023large,huang2022inner}. 
This belief seem to  rest largely on the assumption that verification of correctness should be easier than generation for many reasoning problems--a rather classical argument from computational complexity. There are grounds to be skeptical of this assumption as complexity of the reasoning task should 
be irrelevant to LLM performance if what they are doing is approximate retrieval. 

In this paper, we set out to systematically investigate effectiveness of iterative prompting in the context of {\em Graph Coloring}, a canonical NP-complete reasoning problem. We chose graph coloring as it is representative both of standard classes of reasoning problems studied in AI--propositional satisfiability and constraint satisfaction--and practical problems like scheduling and allocation. Our methodology involves a principled empirical study of the performance of GPT4 on two tasks: solving a large suite of random graph coloring instances and, separately, verifying the correctness of the candidate colorings--both in direct and iterative modes. In iterative modes, we experiment both with an LLM critiquing LLM-produced solutions and an external, guaranteed correct reasoner verifying solutions. In both cases, we analyze whether the content of criticisms actually affects bottom line performance. 

Our results indicate that in direct mode, LLMs are, perhaps not surprisingly, pretty bad at solving graph coloring instances. More interestingly, as we suspected, they are no better at verifying solutions. In iterative modes, given the inability of LLMs to verify solutions, it should come as no surprise that our experiments show that the strategy of LLMs self-critiquing their solutions does not improve over the baseline. It is {\em actually worse} because the system can't recognize a correct coloring and thus merrily passes over fortuitously correct colorings it has generated, ending up with a wrong one!

We next experimented with an iterative strategy where an external coloring verifier does the backprompting. Here we looked at three different types of back prompting: (1) the verifier just asks the LLM to try again when the coloring is incorrect, (2) the verifier gives a backprompt showing the first violated constraint in the current candidate coloring and (3) the verifier sends a backprompt showing all violated coloring constraints. We note that these three strategies do lead to modest improvements in the bottom-line performance--improving from about 16\% to nearly 40\%. The surprising finding however is that the minimal information "try again" feedback is nearly as effective as the ones with meaningful backprompts. This lead us to consider whether the improvement is due to the type of backprompting (as authors who advocate these types of iterative approaches \cite{yao2023react,tree-of-thoughts,shinn_reflexion_2023,ling_deductive_2023, chen2023teaching, madaan2023selfrefine} seem to assume) or because the answer just happens to be in the top-K completions (even if the LLM is itself not cognizant of it). To check this, we experiment with a version of the direct mode where we query the LLM so that it generates more than one potential solution, and have the external verifier pick out any correct solution in the list. The results show that top-k correctness with an external, guaranteed correct verifier is pretty competitive with any iterative backprompting. 

Our investigation thus raises significant grounds to be skeptical about the effectiveness of iterative prompting techniques in general, and those relying on the self-critiquing capabilities of LLMs in particular. In the reminder of the paper, we discuss related work, present our experimental methodology, and then detail the results of our experiments.

\section{Related Work}

As mentioned in the introduction, there has been a large recent body of work investigating the reasoning capabilities of LLMs \cite{shinn_reflexion_2023,tree-of-thoughts, kojima_large_2022}. The studies span different types of reasoning problems--planning \cite{valmeekam_large_2023}, logic and arithmetic \cite{dziri_faith_2023}, or 24 puzzle \cite{tree-of-thoughts}. The conclusions have also been divergent--with some studies highlighting the limitations of LLMs in reasoning\cite{mccoy2023embers, 4cantreason}, and others arguing that iterative prompting of LLMs can improve their ability to reason. For example, \cite{shinn_reflexion_2023} states {\em  we explore this emergent property of self-reflection in LLMs and empirically show that self-reflection is extremely useful to learn complex tasks over a handful of trials.} This paper focuses on understanding these sorts of claims--and especially of the effectiveness of iterative prompting. 
The problem we chose--graph coloring--is a canonical NP-complete reasoning problem well studied in AI and computer science \cite{russell-norvig-text}. It has rich connections to propositional logical reasoning--specifically satisfiability, constraint satisfaction problems, and is also related to practical problems including resource allocation and scheduling. 


\section{Methodology}

\subsection{The Graph Coloring Problem}

Because we are interested in LLMs' self-critique capabilities, we chose \emph{Graph Coloring}, a reasoning domain which is human readable, provides relatively short description and critique lengths, and, most importantly, is very easy to verify and provide feedback for. Though it is difficult to be certain, we also believe that this domain is diverse enough even at low node and edge counts that the instances we examine are very unlikely to be found in the LLM's training data, thus minimizing the risk of model contamination and memorization.

Graph coloring is a a canonical NP-complete reasoning problem that is related to both propositional satisfiability as well as practical problems like scheduling and allocation. It is broad enough to give insights into reasoning more generally, and simple enough to be specified and evaluated by a human or basic pattern matching.

Common graph coloring benchmark sets consist of the sorts of problems that exact solvers struggle on, boasting triple or quadruple digit numbers of nodes and edges\cite{graphcoloringinstances}. Current language models don't have sufficiently large context windows to process these, and—as we'll see later—are unlikely to do well on graphs with over twenty nodes.

Therefore, we built our own dataset. We use \verb|GrinPy|\footnote{https://pypi.org/project/grinpy/} to handle common graph operations. Each graph is constructed using the Erdős–Rényi method (\(p=0.4\)), modified so that any generation that fails to be planar or happens to be isomorphic to a previously generated one is retried. Once a successful candidate is found, it is compiled into the standard DIMACS format\cite{dimacsrutgers}, appended with a comment containing its precalculated chromatic number. 

\begin{figure}[h]
    \centering
    \includegraphics[width=0.8\linewidth]{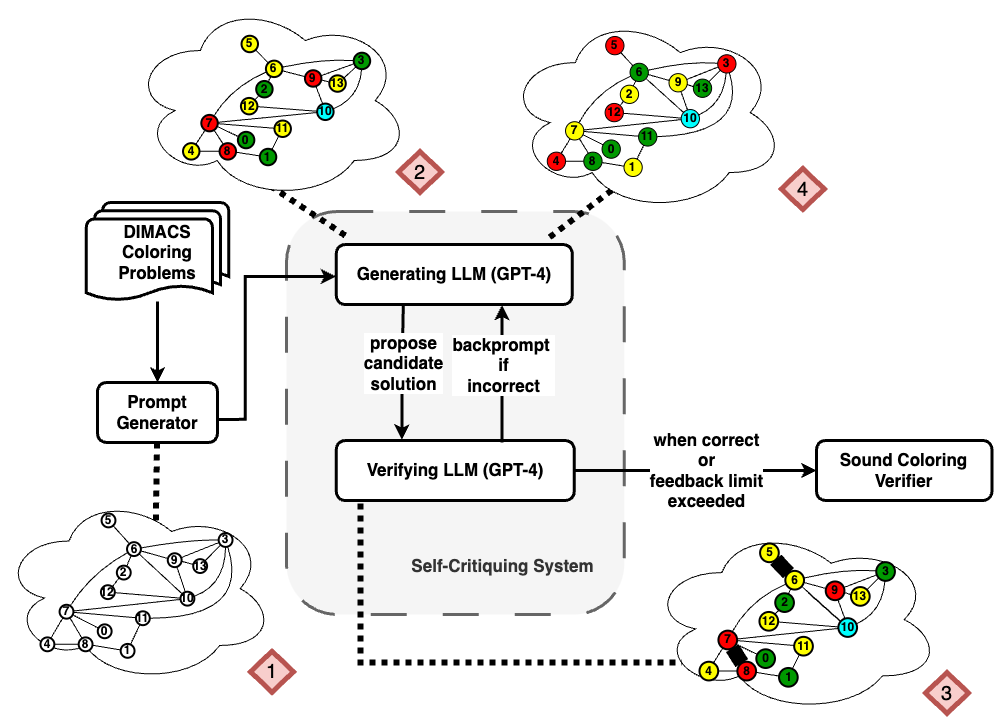}
    \caption{Overview of backprompt architecture for a single instance. Clouds provide an illustrated interpretation of the current state of the problem at different points in the system. Red diamonds indicate progression of a single problem: a planar graph is first passed to GPT-4 acting as a generator (1), which returns a proposed coloring (2). GPT-4 will then be used as a verifier to determine whether the coloring is correct. When not correct, GPT-4 provides feedback, along with previous history, through a backprompt (3) that will be used in the next generation request (4). Each new coloring will be evaluated by the GPT-4 working as a verifier. If GPT-4 determines the coloring to be correct or 15 iterations have passed, it approves the final answer, where it is then evaluated against a sound verifier.}
        \label{fig:pipeline}
\end{figure}

For the following experiments, we generated 100 instances with an average of 24 edges each spread across node counts from 10 to 17—a distribution chosen because empirical probing revealed it to be an area with volatile enough performance to be interesting. An example of one of the graphs we used is shown in Figure \ref{fig:pipeline}, together with the LLM's first response, the backprompt on that response, and the final correct coloring. 

\subsection{Architecture for Iterative Backprompting}


All code and results will be made public. 

\mund{Prompt Generator}

The generator takes a DIMACS instance and constructs a natural language prompt by translating each edge into a sentence and then wrapping the whole in a common set of instructions. We deliberately minimize differences between instances' prompts to reduce how much problem-specific information we leak to the LLM. Examples of each prompt type can be found in the appendix.

\mund{Large Language Model}

Off the shelf, this system allows for the use of any LLM accessible through the OpenAI API: the user need only pass the model name through the appropriate flag at runtime. The present work focuses on GPT-4, the current state of the art, because of recent claims about its "emergent" reasoning capabilities\cite{bubeck2023sparks}.

We provide a system role of "You are a constraint satisfaction solver that solves various CSP problems." and set the temperature to \(0\), thus ensuring output is mostly deterministic.

\mund{Extensibility}

This architecture easily extends to other domains of constraint satisfaction problem solving. In the public repository, we provide a way to add a new domain description by adding just one file to the project in plug-and-play fashion.

\subsection{Backprompt Generation}

In verification mode, the LLM receives a different sort of prompt. Apart from standard instructions, it contains only the graph description and the proposed coloring. It is tasked with verifying correctness, optimality, and whether every vertex has been given an assignment. If the coloring is incorrect, it must reply with a set of contradicting edges.

As a comparison point, we also construct a guaranteed correct verifier, with the ability to list every single contradicting edge. Since LLM responses are also in natural language, we first translate them into a format amenable to analysis. To make this process more consistent, we design our initial prompt to describe an exact output format to which the model conforms. Then, the response is evaluated for correctness.

In both cases, if the verifier says the answer is correct, we end there. If it has been more than 15 rounds (16 total queries), we give up. Otherwise, a backprompt is created, wrapped in standard instructions, appended to the previous message history, and sent back to the model as a new prompt.

In this domain a valid piece of error feedback consists of a pair of vertices which were given the same color but share an edge. To construct a backprompt, we have to decide exactly how much feedback to give. We examine five cases:

\begin{enumerate}
    \item \textbf{None}: A single iteration baseline. No backprompting.
    \item \textbf{Pass/Fail}: The only feedback given is that the answer was incorrect.
    \item \textbf{First}: Only the first error encountered is returned.
    \item \textbf{Full}: A comprehensive list of errors.
    \item \textbf{LLM}: Feedback is provided by the language model through a separate prompt, given in the appendix. We pass any and all response back to the generator, regardless of its validity or correctness.
\end{enumerate}

By comparing results under these regimes, we can deduce how much of the given information the LLM is actually using, versus how much of the performance increase stems from merely getting more tries. We also compare these cases to four further cases: higher temperature, single iteration queries which ask for multiple answers. These do not involve any backprompting, reprompting, or giving any information past the original prompt to the LLM.

\begin{enumerate}
    \item[6-8.] \textbf{Top 5}: With temperatures 0.5, 1, and 1.5, query the LLM for \(n=5\) responses.
    \item[9.] \textbf{Top 15}: With a temperature of 1, query the LLM for \(n=15\) responses.
\end{enumerate}

\subsection{Verification}

In order to gain more insight into their LLM verification, we examine how well they find errors in proposed colorings. Intuitively, these should be very easy to identify: if the two vertices making up an edge share a color, immediately return that edge. Algorithmically, all this requires is looping over edges and comparing each vertex's color to that of its partner.

We use the same pipeline for this analysis, but construct a new domain we call \verb|color_verification|. The LLM is prompted to check correctness, optimality, and if every vertex has been assigned in the coloring. If the coloring is incorrect, it is instructed to list errors in the coloring, that is, if two connected nodes share a color, it is to return the edge to represent the error. No backprompts are given. We use the same graph instances from before, but generate four kinds of colorings to test the model on:

\begin{enumerate}
    \item \textbf{Correct}: Optimal colorings with no errors, generated via iterated, randomized greedy algorithm (with a precomputed chromatic number to ensure optimality)
    \item \textbf{Ablated}: The previous set of colorings, each with a random node changed to one of its neighbor's colors
    \item \textbf{Non-optimal}: The correct set, with a randomly chosen color partially recolored to a new shade
    \item \textbf{Random}: Completely randomly assigned colors, with the number of different colors equal to the graph's chromatic number
    \item \textbf{LLM}: Colorings randomly selected from the LLM-generated outputs of the previous experiment 
\end{enumerate}

\section{Results}

\subsection{Backprompting as Self-Critique}

\begin{table}[t]
  \caption{Summary of Backprompt Techniques}
  \label{techniques}
  \centering
  \begin{tabular}{l>{\ttfamily\raggedright\arraybackslash}p{2.7in}}
    \toprule
    Strategy &
    \textnormal{Example Prompt} \\
    \midrule
    Direct LLM & 
    Color the following graph, described as a set of edges, such that no two vertices on the same edge share a color. You may use at most 3 colors. Vertex 0 is connected to vertex 2...  \\
    \arrayrulecolor{black!30}\midrule
    Iterative: LLM Self-Critique 
    &  This is incorrect. Feedback: Vertices 0 and 11 share an edge and are both colored with Color 1. Vertices 5 and 11 [...] Using this feedback...\\
    \arrayrulecolor{black!30}\midrule
    Iterative (with external Verifier): Pass/Fail & 
    This is not correct. Using the previously provided graph... \\
    \arrayrulecolor{black!30}\midrule
    Iterative (with external Verifier): First error & 
    This is not correct. Vertex 1 and vertex 7 were both colored Color 1 despite being connected by an edge... \\
    \arrayrulecolor{black!30}\midrule
    Iterative (with external Verifier): All errors & 
    This is not correct. Vertex 1 and vertex 7 were both colored Color 1 despite being connected by an edge. Vertex 2 and  vertex 4 were both colored Color 0 despite...\\
    \arrayrulecolor{black}\bottomrule
  \end{tabular}
\end{table}

\begin{figure}[h]
\centering
\begin{minipage}{.5\textwidth}
  \centering
  \includegraphics[width=\linewidth]{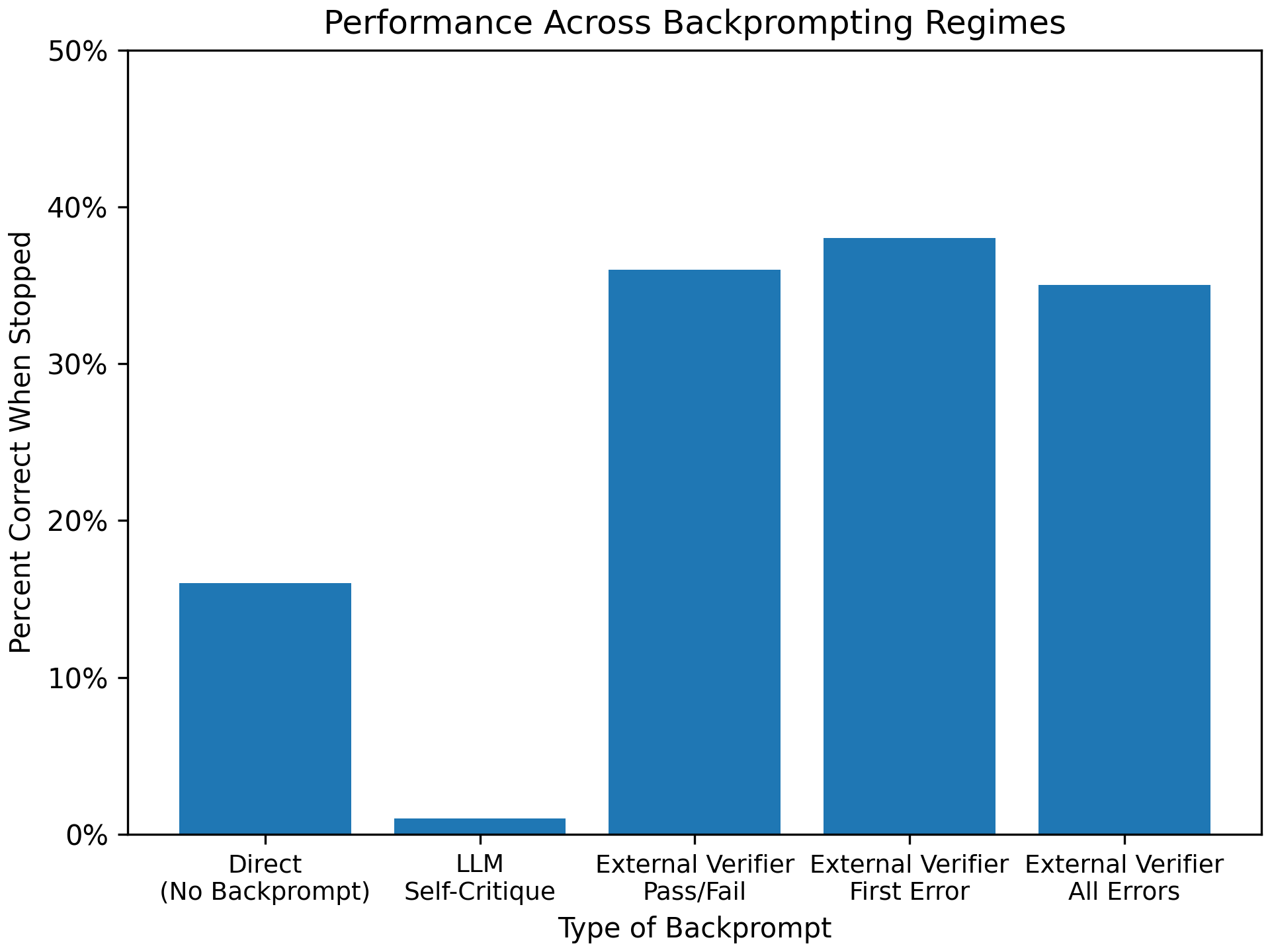}
  \captionof{figure}{Performance versus backprompting technique. Correctness is evaluated for the response the verifier claims as correct, or after 15 iterations.}
  \label{fig:scores}
\end{minipage}%
\begin{minipage}{.5\textwidth}
  \centering
  \includegraphics[width=\linewidth]{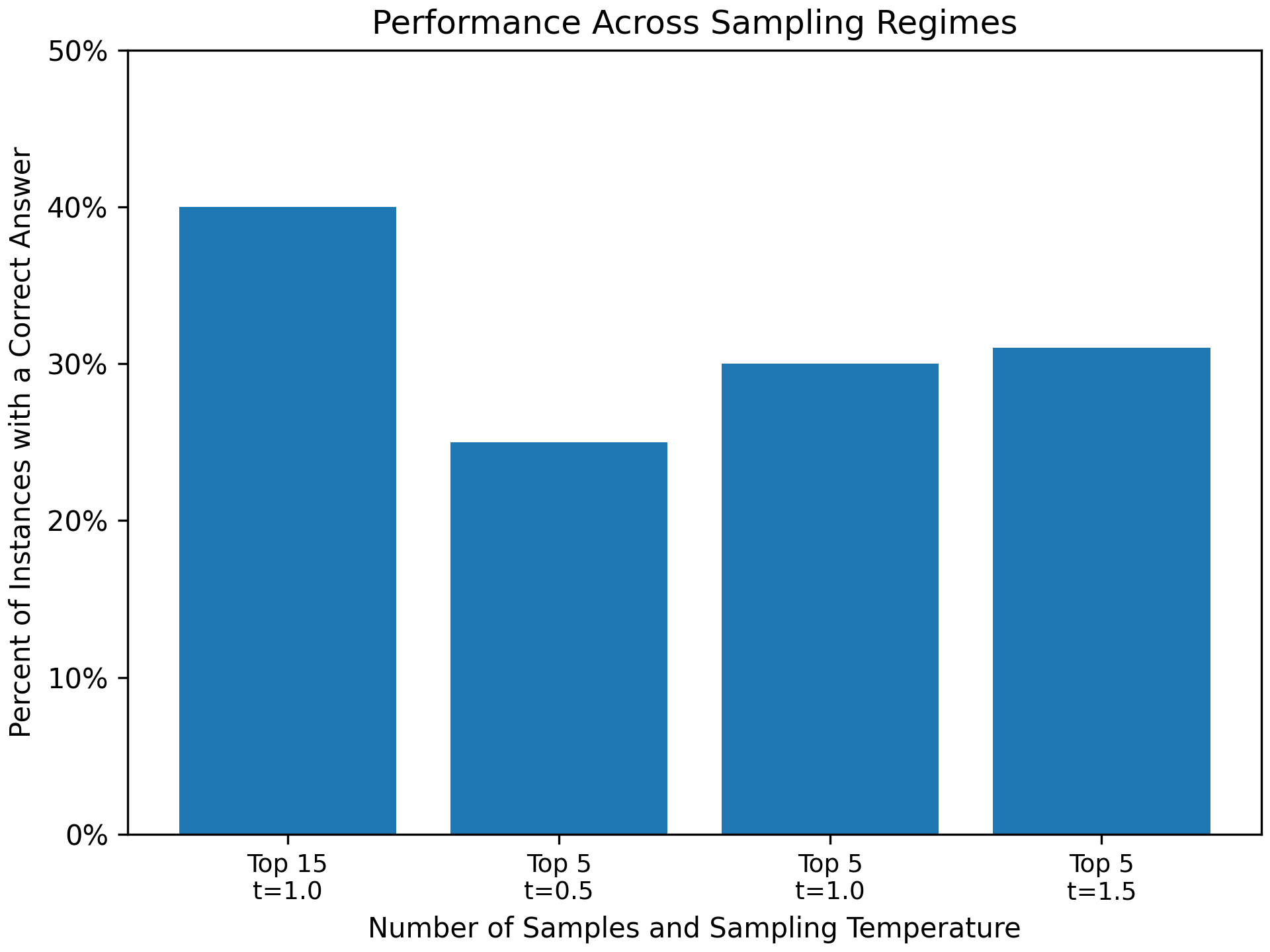}
  \captionof{figure}{Performance versus sampling technique. An instance is marked correct if \emph{any} answer in the top n was correct.}
  \label{fig:topscores}
\end{minipage}
\end{figure}

Prompting the LLM, evaluating the answer, and moving on to the next instance without any backprompts whatsoever gives a baseline score of 16\%. When we run the same instances, but this time backprompt the LLM with feedback generated by the same language model acting as a verifier, performance plummets--only a single instance of the 100 was answered correctly. 

The problem is caused by the lack of an accurate stopping condition. If the system ever outputs a correct coloring during a backprompting session, we expect a verifier to stop it. However, in the self-verification case, the LLM doing the verification can fail to notice success and instead produce spurious feedback. This is exactly what happens. 

At some point in the backprompts of 40 instances, the generating model returned an optimal coloring. In none of those instances did the verifying GPT realize this. In 39 cases, it hallucinated pairs of vertices that it claimed were adjacent and same-colored. In the one case marked correct, the coloring was provided after the final backprompt, and so became the model's final answer by virtue of timeout. This also points to the model's hesitancy to agree that a coloring is correct. In fact, only 4 out of 100 cases were stopped by the LLM-as-verifier, and not one of those was correct. Whether bad feedback itself is worsening the results, or it's merely the case that correct responses tend to be earlier in the backprompt sequence--optimistically viewed as a result of being higher probability completions which are ruined by a self-destructive thinking process--is unclear. Our results here and in the next few subsections are so far conflicting.

The results when backprompted with a sound verifier seem, at first, a lot more promising. The number of instances correctly answered nears 40\%, but if this is supposed to indicate that GPT-4 is listening to, improving with, and reasoning from feedback, then we should expect more informative and accurate backprompts to yield better results. However, in this domain, the raw scores (see Figure \ref{fig:scores}) don't bear this out. When run with a sound verifier, the differences between binary feedback, a single error, or the full suite of mistakes are insignificant.

We can relax our analysis of the LLM self-critique case by labeling an instance as correct if at \emph{any} point during the backprompt chain, the LLM generated a correct coloring. This is equivalent to rerunning the experiment with a combined feedback system: the sound verifier is in charge of stopping while allowing the LLM to write all the (still potentially spurious) feedback. Given this modification, it scores a comparable 40\%. Using this charitable number, all four types of backprompting give roughly similar results.

It seems then that feedback or lack thereof is thus less important to the improvement of the score than number of iterations: if the model has fifteen chances to generate a correct answer, it is much more likely to succeed. We test this idea by querying the same set of 100 instances, but now allowing for higher temperatures and receiving multiple, separate, non-interacting responses. The results make up the rest of Figure \ref{fig:topscores}. With \verb|n=5|, it's close, not quite there, but with \verb|n=15| (\verb|t=1.0|), the performance is comparable to backprompting, achieving a score of 40\%. 

In other words: blindfolded guessing does just as well as careful, crafted feedback.

The rest of our analysis examines where the system is going wrong. We will attempt to answer two questions: to what extent is the LLM capable of determining if a solution is right or wrong? How, if at all, does the LLM respond to feedback?

\subsection{Verification by LLM}

We test GPT-4's ability to verify colorings on the same instances, but we generate five different kinds of colorings for each. What is immediately obvious is a result that exactly agrees with the LLM self-verification results above: the model is unwilling to mark almost \emph{any} answer as correct. Out of 100 optimal colorings, it only agreed that 2 were correct. Expanding to the entire set of 500 colorings, of which 118 of them are correct, it only claimed 30 of them as correct. Of those, it was right 5 times. This isn't because of any special property of correctness--the same holds true in the non-optimal coloring set, in which it only marked 10\% of instances as non-optimal. 

Overall, this pattern holds. Fewer than ten percent of cases resulted in a "correct", "non-optimal", or "missing assignment" response from the LLM. Among those, the behavior looks somewhat random. In around a quarter of instances, it responds with a "this is incorrect" verification where the explanation matches reality, and it only manages this by naming no more than a single edge, which minimizes the chance of misstating something.

\begin{table}[]
\caption{Distribution of hallucinations during verification task. This table counts the number of instances that featured each type of hallucination and compares it to the total number of erroneous edges encountered across all coloring instances in each subset.}
  \label{table:hallucination}
  \centering
\begin{tabular}{@{}lllllll@{}}
\toprule
                                 & \multicolumn{4}{c}{Hallucinations}               & \multicolumn{2}{c}{Coloring} \\ \midrule
\multicolumn{1}{l|}{}            & Vertex & Edge & Both & \multicolumn{1}{l|}{None} & Errors       & Correct       \\
\multicolumn{1}{l|}{Correct}     & 29     & 72   & 7    & \multicolumn{1}{l|}{2}    & 0            & 100           \\
\multicolumn{1}{l|}{Ablated}     & 24     & 52   & 5    & \multicolumn{1}{l|}{24}   & 187          & 0             \\
\multicolumn{1}{l|}{Non-optimal} & 18     & 65   & 3    & \multicolumn{1}{l|}{10}   & 0            & 0             \\
\multicolumn{1}{l|}{Random}      & 10     & 26   & 5    & \multicolumn{1}{l|}{66}   & 736          & 0             \\
\multicolumn{1}{l|}{LLM}         & 26     & 41   & 6    & \multicolumn{1}{l|}{27}   & 240          & 18            \\ \midrule
\multicolumn{1}{l|}{Total}       & 107    & 256  & 26   & \multicolumn{1}{l|}{129}  & 282          & 118
\end{tabular}
\end{table}

Table \ref{table:hallucination} summarizes the results. Note that, proportionally, hallucinations decrease when the error rate of the domain increases. That is to say, when there are more incorrect edges, the model is more likely to point to one of them. Intuitively, this makes sense: it's easier to guess one edge which is wrong when half of all the edges are miscolored, as is the case on average among randomly colored graphs.

Edge hallucinations are more common than vertex. Essentially, typical behavior is to pick two vertices that are the same color in the coloring, but which aren't associated by an edge in the graph description, and claim that they are connected and thus illegally colored. Vertex color hallucination is when the reverse happens: instead of ascribing an edge to same-color nodes, the colorings of two connected vertices are misstated. The overlap between the two cases, where a non-existent edge is declared to be violated by non-existent colorings is much rarer than either. Note that it never hallucinates new vertex \emph{names}, only that vertices which are in graph have colors differing from reality.

Even rarer cases did spring up in the response data. At times the model lost track of the question being asked and reversed it, explicitly claiming that two same-colored vertices violate the conditions because they \emph{aren't} connected; or it began to contradict itself mid-deduction, making multiple claims about a vertex's color. We relegate these examples to the appendix.

Our overall conclusion is that, despite the common-sense nature of this domain, the LLM's verification powers are surprisingly weak.

\subsection{Inside the Backprompt Chain}

To figure out what information it is or isn't using, we examine the evolution of GPT-4's responses within a backprompt chain. We compare three types of informative backprompting: providing the first wrong edge, listing all wrong edges, and choosing a random correct edge to claim is incorrect. The first two cases were described in more detail above. The final one, the so-called "evil" case is new, and provided as a way to check how blindly the system follows corrective advice.

Given a backprompt, we examine the response to it. We only look at the rates of local error correction. Given a backprompt, we consider it "listened to" if the edges it listed as incorrect were changed in the response so that each vertex is a different color from the other. We summarize the results by averaging over backprompts. The results are summarized in Table \ref{table:corrections}.

\begin{table}[h]
\caption{Local error correction rates per backprompt information type. Full (any) gives credit if any edge mentioned in the backprompt was corrected. Full (all) gives each backprompt response a percentage score calculated from the number of mentioned edges which were corrected divided by the total number of edges mentioned in the backprompt. Evil backprompting claims a random correct edge is incorrect.}
  \label{table:corrections}
  \centering
\begin{tabular}{@{}llll@{}}
\toprule
           & \# Backprompts & \# Incorrect Edges Fixed & \% Incorrect Edges Fixed \\ \midrule
First      & 1066           & 1004                     & 94\%           \\
Full (any) & 1102           & 1077                     & 98\%           \\
Full (all) & 1102           & 2870                     & 84\%           \\
Evil       & 1083           & 1017                     & 94\%           \\ \bottomrule
\end{tabular}
\end{table}

Even though performance was unaffected, GPT \emph{did} correct most errors that were pointed out. However, it didn't discriminate between real errors or the evil case's false ones, blindly applying local "fixes" without regard for overall correctness. 

\section{Conclusion}

In this work, we have set out to investigate the effectiveness of iterative prompting strategies in improving the accuracy of LLMs on reasoning problems. We were motivated, in particular, by claims in prior work that even when LLMs produce incorrect answers at first, they are good at self-critiquing and improving their answers. Our results on graph coloring call these claims into question. They show that LLMs are in fact very poor at verifying solutions (in our case, colorings), something that is critical for self-critiquing. Not surprisingly, iterative framework with LLMs self-critiquing does even worse than LLMs directly generating a single answer. We do show that iterative prompting can help when there is an external provably correct verifier in the loop. Even here, we found that the actual content of iterative back prompts is not important, and that the improvements seen can also be obtained by just having the LLM produce multiple answers, and letting verifier check and pick any correct answer that was fortuitously generated. Our results thus raise legitimate questions about claims of the effectiveness of iterative prompting, adding further fuel to the skepticism surrounding the reasoning capabilities of LLMs. 

\begin{ack}
Acknowledgements
\end{ack}

\bibliographystyle{plain}
\bibliography{refs}

\appendix
\section{Appendix}

\subsection{Prompts}
All of following examples are built on the same graph instance.
\mytcbinput{appendix/dimacs_example.tex}{DIMACS Format For Graphs}{0}{example:dimacs}
\mytcbinput{appendix/base_prompt.tex}{Baseline{,} Direct Prompt}{0}{prompt:baseline}
\mytcbinput{appendix/example_response.tex}{Example LLM Response}{0}{example:response}
\mytcbinput{appendix/llm_v_prompt.tex}{Prompt To Elicit Verification}{0}{prompt:verification}

\subsection{Backprompts}
\mytcbinput{appendix/llm-self_example_backprompt.tex}{Backprompt Generated From Self-Critique}{0}{backprompt:self}
\mytcbinput{appendix/passfail_example_backprompt.tex}{Backprompt Generated From Sound Verifier (Pass/Fail)}{0}{backprompt:passfail}
\mytcbinput{appendix/first_example_backprompt.tex}{Backprompt Generated From Sound Verifier (First)}{0}{backprompt:first}
\mytcbinput{appendix/full_example_backprompt.tex}{Backprompt Generated From Sound Verifier (Full)}{0}{backprompt:full}

\subsection{Complete Example of Iterative Prompt Sequence}
\mytcbinput{appendix/full_trace_example.tex}{Full Sequence of 15 (LLM-Generated) Backprompts}{0}{example:trace}

\subsection{LLM as Verifier}
In this section, multiple equals signs in a row ("===") are dividers between separate examples, not parts of the examples.
\mytcbinput{appendix/LLM_V_example_normal.tex}{Examples of LLM Output on the Verification Task}{0}{example:normal_v_responses}
\mytcbinput{appendix/LLM_V_example_weird.tex}{Examples of (Rare) Mangled{,} Illogical{,} and Otherwise Strange Output on the Verification Task}{0}{example:weird_v_responses}



\end{document}